\documentclass[10pt,twocolumn,letterpaper]{article}

\usepackage{iccv}              % To produce the CAMERA-READY version
\usepackage[accsupp]{axessibility}

\usepackage{overpic}
\usepackage{adjustbox}
\usepackage{multirow}
\usepackage{colortbl}
\usepackage{xcolor}

\definecolor{iccvblue}{rgb}{0.21,0.49,0.74}
\usepackage[pagebackref,breaklinks,colorlinks,allcolors=iccvblue]{hyperref}
\usepackage{amsmath}
\usepackage{float}

\newcommand{\blockcomment}[1]{}
\makeatletter
\def\blfootnote{\xdef\@thefnmark{}\@footnotetext}
\makeatother

\def\confName{ICCV}
\def\confYear{2025}

\title{Bridging Continuous and Discrete Tokens for Autoregressive Visual Generation}

\author{
  \textbf{Yuqing Wang}\textsuperscript{1} 
  \quad \textbf{Zhijie Lin}\textsuperscript{2} 
  \quad \textbf{Yao Teng}\textsuperscript{1} \\
  \quad \textbf{Yuanzhi Zhu}\textsuperscript{3} 
  \quad \textbf{Shuhuai Ren}\textsuperscript{4} 
  \quad \textbf{Jiashi Feng}\textsuperscript{2}  
  \quad \textbf{Xihui Liu}\textsuperscript{1}\footnotemark[1]  \\ 
\textsuperscript{1}University of Hong Kong  \quad
\textsuperscript{2}ByteDance Seed \quad
\textsuperscript{3}École Polytechnique\quad
\textsuperscript{4}Peking University \\
}

\usepackage{amsmath,amsfonts,bm}
\usepackage{makecell}
\def\eqref#1{equation~\ref{#1}}
\def\1{\bm{1}}

\def\vq{{\bm{q}}}

\def\vx{{\bm{x}}}

\def\vz{{\bm{z}}}

\def\mX{{\bm{X}}}

\DeclareMathAlphabet{\mathsfit}{\encodingdefault}{\sfdefault}{m}{sl}
\SetMathAlphabet{\mathsfit}{bold}{\encodingdefault}{\sfdefault}{bx}{n}

\begin{document}
\maketitle
\blfootnote{\quad $*$Corresponding author.}

\begin{abstract}
Autoregressive visual generation models typically rely on tokenizers to compress images into tokens that can be predicted sequentially. A fundamental dilemma exists in token representation: discrete tokens enable straightforward modeling with standard cross-entropy loss, but suffer from information loss and tokenizer training instability; continuous tokens better preserve visual details, but require complex distribution modeling, complicating the generation pipeline. In this paper, we propose TokenBridge, which bridges this gap by maintaining the strong representation capacity of continuous tokens while preserving the modeling simplicity of discrete tokens. To achieve this, we decouple discretization from the tokenizer training process through post-training quantization that directly obtains discrete tokens from continuous representations. Specifically, we introduce a dimension-wise quantization strategy that independently discretizes each feature dimension, paired with a lightweight autoregressive prediction mechanism that efficiently model the resulting large token space. Extensive experiments show that our approach achieves reconstruction and generation quality on par with continuous methods while using standard categorical prediction. This work demonstrates that bridging discrete and continuous paradigms can effectively harness the strengths of both approaches, providing a promising direction for high-quality visual generation with simple autoregressive modeling. Project page: \url{https://yuqingwang1029.github.io/TokenBridge}.
\end{abstract}    
\section{Introduction}
\label{sec:intro}

\begin{figure}[!tbp]
 % \vspace{-2pt}
  \centering
  \includegraphics[width=0.78\linewidth]{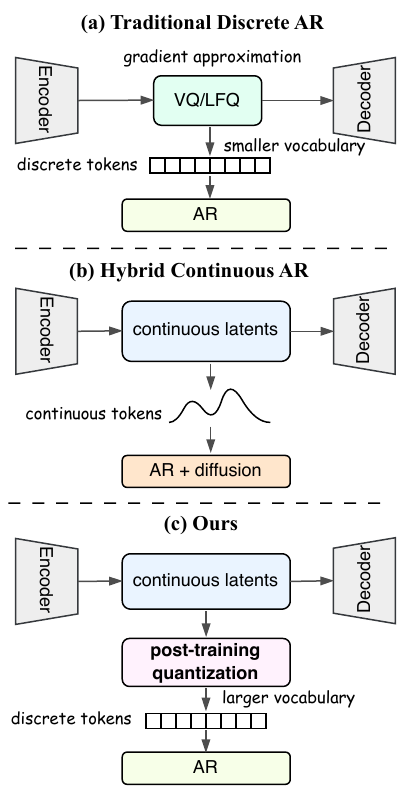}
  \vspace{-3pt}
    \caption{\textbf{Comparison of different autoregressive visual generation approaches.} (a) Traditional discrete tokenization incorporate quantization during training, resulting in tokenizer training instability and limited vocabulary size that restricts representational capacity. (b) Hybrid continuous AR models preserve rich visual information but need complex distribution modeling (diffusion or GMM) beyond standard categorical prediction. (c) Our approach bridges these paradigms by applying post-training quantization to pretrained continuous features, maintaining the high representational capacity of continuous tokens while enabling simple autoregressive modeling.}
  \label{fig:compare}
   \vspace{-15pt}
\end{figure}

\begin{figure*}[!tbp]
  \centering
  \includegraphics[width=\linewidth]{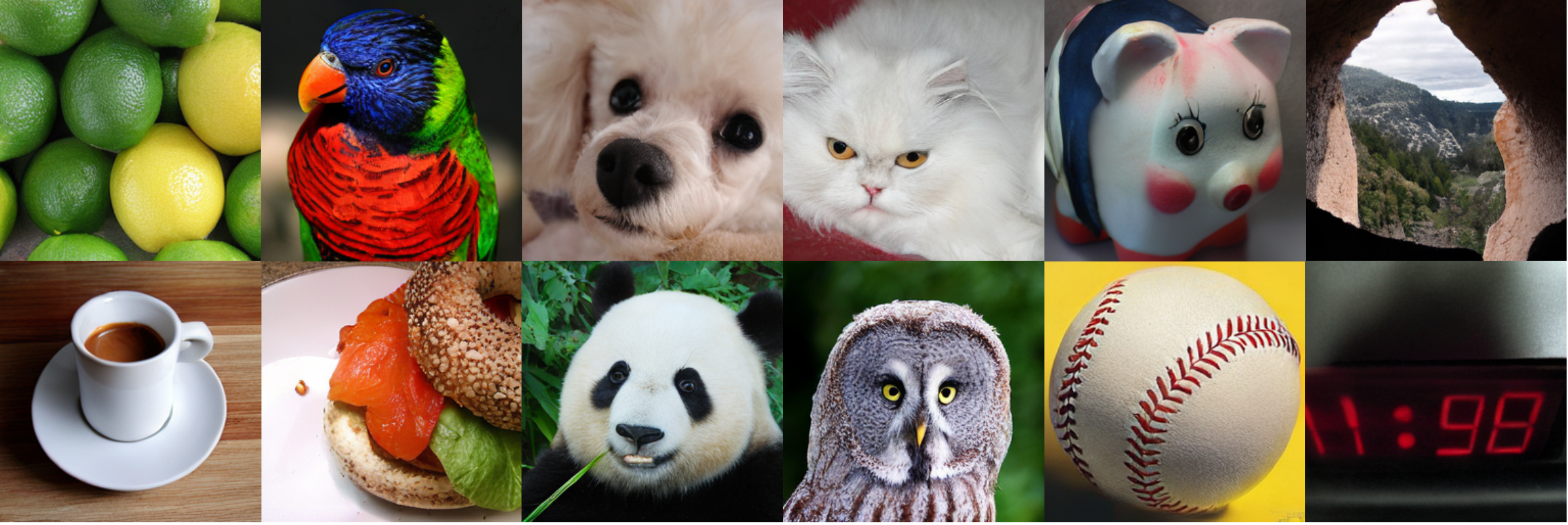}
  \vspace{-20pt}
 \caption{\textbf{Generated samples from TokenBridge.} Class-conditional generation results on ImageNet~\cite{imagenet} 256×256 demonstrating fine details and textures across diverse categories including animals, food, objects, and scenes.}
  \label{fig:vis}
   \vspace{-15pt}
\end{figure*}

Autoregressive visual generation models~\cite{chen2020generative,sun2024autoregressive,tian2024visual,yu2023language,chen2024next,wang2024parallelized,pang2024randar,kondratyuk2023videopoet,wang2024loong,villegas2022phenaki,wu2022nuwa} have emerged as a promising paradigm for visual synthesis, inspired by the way language models predict text tokens. These models rely on image tokenizers~\cite{vae,esser2020taming,vqvae,vit-vqgan,yu2023magvit,bsq} that convert image content into discrete or continuous tokens, which are then predicted sequentially through autoregressive modeling in a next-token prediction paradigm. This analogous modeling approach enables flexible integration with text tokens for multimodal tasks~\cite{team2024chameleon,wang2024emu3}, and allows them to benefit from architectural innovations and scaling techniques established in LLM research~\cite{henighan2020scaling,radford2019language,gpt3}. Despite these advantages, a fundamental dilemma remains in the choice between discrete and continuous token representations, which significantly impacts model complexity and generation quality.

%To better understand this dilemma, we examine the two main categories of existing approaches. 
Traditional discrete tokenization methods~\cite{vqvae,esser2020taming,mentzer2023finite,bsq,yu2023language} employ various quantization techniques, such as VQ~\cite{esser2020taming} and LFQ~\cite{yu2023language},  to map continuous features into discrete tokens during training. However, these methods face two key issues. First, quantization is inherently non-differentiable, requiring gradient approximations that  introduce  complicated optimization and  training instability~\cite{Zheng_2023_CVQ,vit-vqgan}. Second, discrete tokenizers struggle with a fundamental trade-off: limited vocabularies cannot fully capture fine visual details, while larger vocabularies often suffer from poor codebook utilization and increase modeling complexity~\cite{openmagvit2,zhu2024scaling}.

The alternative approach employs continuous VAE-based tokenizers~\cite{vae}, which preserve rich visual information by optimizing latent representations through direct gradient back-propagation. However, continuous latent tokens cannot be directly modeled by standard autoregressive approaches that rely on categorical prediction, forcing recent methods~\cite{givt,mar} to replace the classification objective with specialized distribution modeling techniques, which necessitates additional model components (e.g., diffusion head~\cite{DDPM_paper} or Gaussian Mixture Models~\cite{reynolds2009gaussian}). While effective, these approaches increase the complexity of the modeling pipeline and often require more sophisticated training and sampling procedures. 

In this work, we explore bridging continuous and discrete token representations for autoregressive visual generation.  
The fundamental challenge lies in how to preserve the strong representation capacity of continuous tokens while maintaining the modeling simplicity of discrete tokens. This is difficult to achieve under the conventional paradigm. We propose to decouple the discretization from the tokenizer training by applying training-free quantization after continuous tokenizer training, which makes it easier to achieve the above two objectives: (1) maintaining capacity can be achieved through fine-grained quantization of pretrained high-quality token representations, and (2) modeling simplicity is naturally guaranteed due to the discrete nature of the resulting tokens, which enables standard categorical prediction.

Specifically, we apply a dimension-wise quantization strategy that independently discretizes each feature dimension of pre-trained continuous VAE~\cite{vae} features. This approach circumvents the optimization instabilities inherent in training-based discrete tokenizers while allowing flexible selection of vocabulary size, and aligning well with pretrained continuous latent space. By preserving the distribution characteristics of the pretrained features, our method effectively maintains the rich visual information without requiring massive explicit codebooks. However, this fine-grained discretization results in an exponentially large token space that poses computational challenges for direct classification. To address this, we propose a lightweight autoregressive mechanism that decomposes token prediction into a series of dimension-wise predictions. This approach efficiently captures the critical inter-dimensional dependencies necessary for high-quality generation while avoiding the computational burden of standard classification approaches over large vocabularies.

Extensive experiments demonstrate the effectiveness of our approach. For tokenization, our discrete tokenizer achieves reconstruction quality comparable to continuous VAEs, validating the effectiveness of our post-training quantization. For generation, our model matches the visual quality of continuous approaches while achieving state-of-the-art results on the ImageNet~\cite{imagenet} 256×256 benchmark. To effectively model the exponentially large vocabulary space, our dimension-wise autoregressive prediction significantly outperforms parallel prediction approaches, confirming the importance of autoregressive factorization for capturing complex inter-dimensional dependencies. Additionally, our discrete tokens naturally enable confidence-guided generation capabilities not available in continuous approaches, opening potential for flexible control during the generation process.

 Fig.~\ref{fig:compare} illustrates our approach in comparison with existing methods. Our main contributions can be summarized as follows:

\begin{itemize}

     \item We propose \textit{TokenBridge}, a novel approach that bridges continuous and discrete token representations, demonstrating that standard autoregressive modeling with cross-entropy loss can achieve visual quality comparable to continuous methods while maintaining the simplicity of discrete approaches.

     \item We introduce post-training quantization that directly discretizes pretrained VAE features, eliminating the optimization instabilities of training-based discete tokenizers while preserving the high visual fidelity of continuous representations.

     \item We develop a dimension-wise quantization and prediction strategy that efficiently handles exponentially large vocabulary spaces in two ways: eliminating the need for massive explicit codebooks during tokenization, and making autoregressive prediction over such large spaces computationally feasible.

\end{itemize}

\section{Related Work}
\label{sec:related}

\begin{figure*}[!t]
  \centering
  \vspace{-6pt}
  \includegraphics[width=.8\linewidth]{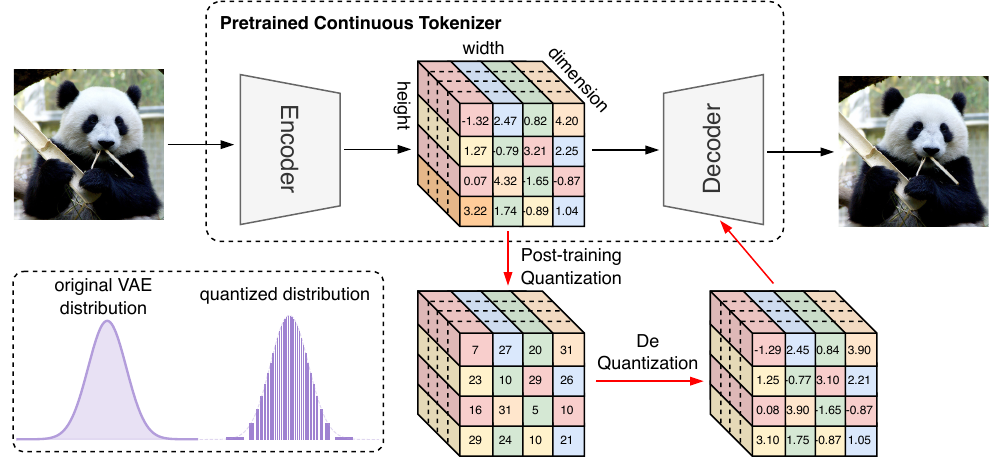}
  \vspace{-6pt}
  \caption{\textbf{Illustration of our post-training quantization process.} The top row shows the pretrained continuous VAE tokenizer, mapping an input image to continuous latent features $\mX \in \mathbb{R}^{H\times W\times C}$ and reconstructing it through the decoder. Our post-training quantization process (middle) transforms these continuous features into discrete tokens by independently quantizing each channel dimension. The bottom-left shows how our approach preserves the original Gaussian-like distribution (purple curve) in discretized form (purple histogram). The right portion demonstrates the de-quantization process that maps indices back to continuous values for decoding.}
  \label{fig:quantize}
   \vspace{-16pt}
\end{figure*}

\noindent\textbf{Visual Tokenization.}
Visual tokenization aims to convert images into tokens for reconstruction and generation, with approaches broadly categorized as continuous or discrete. Variational autoencoders (VAE)\cite{vae} use regularized continuous latent spaces that preserve high visual fidelity, becoming standard in diffusion models\cite{stable_diff, DDPM_paper,peebles2023scalable,DDIM_paper,imagen}. Discrete approaches like VQ-VAE~\cite{vqvae} and VQGAN~\cite{esser2020taming} enable straightforward autoregressive modeling but suffer from codebook collapse and information loss. Although recent methods such as LFQ~\cite{yu2023language}, FSQ~\cite{mentzer2023finite} and BSQ~\cite{bsq} reduce training instability and improve codebook size, their large codebooks present challenges for downstream modeling and require more powerful generation capabilities to achieve high-quality results. Our post-training quantization bridges continuous and discrete tokens by directly quantizing pretrained continuous features, leveraging pretrained VAEs for high-fidelity representation while maintaining the benefits of discrete tokens for autoregressive modeling.

\noindent\textbf{Autoregressive Image Generation.}
Autoregressive image generation has evolved from computationally expensive pixel-based approaches like PixelRNN/CNN~\cite{van2016pixel,salimans2017pixelcnn++} to more efficient token-based methods. Typical models built on discrete tokens include DALL-E~\cite{ramesh2021dalle} and VQGAN~\cite{esser2020taming,vit-vqgan,razavi2019generating,Zheng_2023_CVQ,rq}, while MaskGIT~\cite{chang2022maskgit} and VAR~\cite{tian2024visual} further improved efficiency through masked modeling and multi-scale approaches.
To address quality limitations, hybrid methods like GIVT~\cite{givt} employ Gaussian Mixture Models to predict continuous tokens, while MAR~\cite{mar} introduces diffusion-based token prediction. However, these approaches complicate the model architecture and generation pipeline, requiring additional components beyond standard autoregressive modeling. Infinity~\cite{infinity} takes a different approach with bitwise modeling and self-correction mechanisms. 
In contrast, our approach can maintain autoregressive simplicity while achieving continuous-quality generation without specialized training procedures or elaborate correction schemes.

\section{Method}
\label{sec:method}
Our approach, termed TokenBridge, addresses the fundamental tension between discrete tokens{\textquotesingle} modeling simplicity and continuous tokens{\textquotesingle} representational capacity by reversing the conventional paradigm: instead of quantizing during tokenizer training, we apply feature quantization after a continuous tokenizer has been fully trained. This post-training approach enables us to preserve the rich visual information captured by continuous representations while gaining the modeling benefits of discrete tokens.

The TokenBridge framework comprises two primary components: (1) a post-training dimension-wise quantization strategy that transforms pretrained continuous VAE features into discrete tokens without degrading their reconstruction capacity (Sec.~\ref{sec:3.1}), and (2) an efficient autoregressive prediction mechanism that handles the resulting large vocabulary space by decomposing token prediction into a sequence of dimension-wise predictions (Sec.~\ref{sec:3.2}). Finally, we detail the training and inference procedures for the autoregressive generation model that operates on these discrete tokens (Sec.~\ref{sec:3.3}).

% \subsection{Continuous-to-Discrete Tokenization}

\subsection{Post-Training Quantization}
\label{sec:3.1}

As shown in Fig.~\ref{fig:quantize} (top row), our method starts with continuous latent features $\mX \in \mathbb{R}^{H\times W\times C}$ extracted by a pretrained VAE encoder. Our goal is to discretize these continuous features while preserving their rich visual information. A natural consideration would be to apply Vector Quantization methods~\cite{vqvae, esser2020taming}. However, these approaches would be ineffective in our context, as achieving near-lossless compression would require an impractically large codebook (exponential in feature dimensions), making storage and lookup computationally prohibitive.

Instead, we implement training-free dimension-wise quantization that operates on each channel independently (Fig.~\ref{fig:quantize}, middle). By quantizing each dimension separately rather than entire vectors, we efficiently overcome the codebook size limitation while enabling much finer quantization granularity per dimension. Our dimension-wise approach effectively leverages two key properties of VAE features: (1) their bounded nature due to KL constraints results in a finite value range, allowing effective quantization with limited discrete levels across all feature dimensions; and (2) their near Gaussian distribution allows for efficient non-uniform quantization that allocates more quantization levels to frequently occurring values, as illustrated in Fig.~\ref{fig:quantize} (bottom-left), where the discretized form (purple histogram) preserves the characteristics of the original distribution (purple curve).

\vspace{3pt}
\noindent \textbf{Quantization.}
For a feature vector $\vx = \big( x^1, x^2, \cdots, x^C \big) $ from the feature map $\mX$, we \textit{first} normalize each dimension to facilitate quantization based on Gaussian properties. We identified practical bounds $[\alpha_{\text{min}},\alpha_{\text{max}}]$ from our experiments, then map these to $[-r,r]$ (where $r=3$ corresponds to three standard deviations):
\begin{equation}
    \hat{x}^c = \text{clip}( 2r \cdot \frac{x^c - \alpha_{\text{min}}}{\alpha_{\text{max}} - \alpha_{\text{min}}} - r,~ -r,~ r) ,
\label{eq:norm-feat}
\end{equation}
where $\text{clip}(a, b, c)$ constrains input $a$ to lie between bounds $b$ and $c$. This normalization preserves the relative distribution while enabling Gaussian-based quantization.

\textit{Next}, we establish quantization boundaries $\{b_i\}_{i=0}^{B}$ by dividing the standard normal distribution into $B$ regions of equal probability:
\begin{equation}
    \Phi(b_{i+1}) - \Phi(b_i) = \frac{1}{B}, \quad i \in \{0,\ldots,B-1\},
\label{eq:non-uniform-quantization}
\end{equation}
where $\Phi(\cdot)$ represents the cumulative distribution function. This non-uniform approach allocates more quantization levels to higher-probability regions, efficiently utilizing limited quantization resources.

For each interval $[b_i, b_{i+1}]$, we compute a reconstruction value as the expected value within that range:
\begin{equation}
\gamma_i = \mathbb{E}[\xi|b_i \leq \xi < b_{i+1}], \quad \xi \sim \mathcal{N}(0,1) .
\end{equation}

\textit{Finally}, we determine the quantization index $q^c$ for each normalized value $\hat{x}^c$ by finding the closest reconstruction value:
\begin{equation}
    q^c = \mathop{\arg\min}_{ 0 \leq i < B} | \gamma_i - \hat{x}^c | .
\label{eq:quanti_x}
\end{equation}
This process transforms continuous features into discrete tokens while maintaining essential distribution characteristics, enabling standard categorical prediction.

\vspace{3pt}
\noindent \textbf{De-quantization.} 
Since our autoregressive model predicts discrete indices while the VAE decoder requires continuous features, de-quantization is necessary for image generation. Each quantization index $q^c$ is mapped to its corresponding reconstruction value $\gamma_{q^c}$, then transformed back to the original feature range:
\begin{equation}
x^c = \frac{(\gamma_{q^c} + r)}{2r} \cdot (\alpha_{\text{max}} - \alpha_{\text{min}}) + \alpha_{\text{min}} .
\end{equation}
This process enables direct use of the pretrained VAE decoder with minimal performance degradation. The complete de-quantization process is shown in the right portion of Fig.~\ref{fig:quantize}.

Although our method is motivated by Gaussian distribution properties, we found that linear quantization with sufficient granularity also performs well with only slightly lower performance, demonstrating our approach's robustness to different post-training quantization schemes.

\begin{figure}[!t]
  \centering
  \includegraphics[width=.9\linewidth]{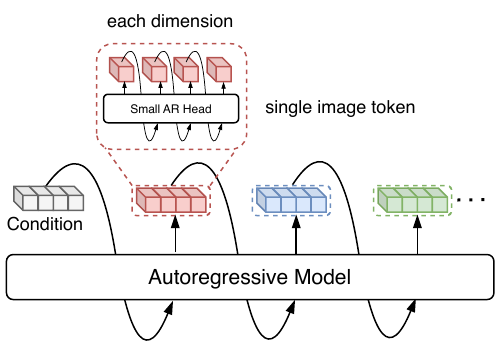}
  \vspace{-10pt}
  \caption{\textbf{Our autoregressive generation process.} At the spatial level, our model autoregressively generates tokens conditioning on previous positions. For each spatial location (highlighted in pink), we apply dimension-wise sequential prediction to efficiently handle the large token space. This approach decomposes the modeling of each token into a series of smaller classification problems while preserving essential inter-dimensional dependencies.}
  \label{fig:ar}
  \vspace{-13pt}
\end{figure}

\subsection{Efficient Large-Vocabulary Token Modeling}
\label{sec:3.2}

While our post-training dimension-wise quantization effectively preserves the representational capacity of continuous tokens, it introduces a computational challenge: an exponentially large token space, comprising $B^C$ possible combinations per spatial location. This makes direct classification through standard softmax computationally infeasible. A straightforward approach would be to independently model and classify each dimension, but our experiments reveal significant interdependencies across the channel dimension that are crucial for high-quality image generation, making such parallel independent prediction impractical.

To address this problem, we introduce a lightweight autoregressive head designed for dimension(channel)-wise next-token prediction at each spatial location, as depicted in Fig.~\ref{fig:ar}. Specifically, for the quantization index vector $\vq=\big( q^1, \cdots , q^C \big)$ at a given spatial location, we model their joint distribution $p(\vq)$ across the channel dimension:
\begin{equation}
  p(\vq) = \prod_{c=1}^C p(q^c|\vq^{<c}, \vz)
\label{eq:ar_head}
\end{equation}
where $q^c$ represents the quantized value for channel $c$, $\vq^{<c}$ denotes all quantized values from preceding channels, and $\vz$ represents the context features from the spatial autoregressive backbone (detailed in~\cref{sec:3.3}).

This autoregressive head predicts a distribution over $B$ possible values for each channel, conditioned on previously generated tokens and context features. By decomposing token prediction into a series of smaller classification problems, this approach makes modeling the exponentially large vocabulary space computationally feasible, and preserving critical inter-channel dependencies.

To further enhance generation quality, we optimize the dimension generation order based on frequency characteristics, prioritizing those carrying more low-frequency information. By analyzing the spectral properties of each dimension through Fast Fourier Transform (FFT)~\cite{nussbaumer1982fast}, we sort dimensions according to their proportion of low-frequency energy. This approach ensures that structural information is generated before fine details, which empirically outperforms the default sequential arrangement.

\subsection{Autoregressive Generation Framework}
\label{sec:3.3}

As illustrated in Fig.~\ref{fig:ar}, our framework integrates spatial autoregressive generation with dimension-wise token prediction. The joint probability distribution across all spatial locations and channels is expressed as:
\begin{equation}
\begin{aligned}
p(\vq) &= \prod_{h,w} \prod_{c=1}^C p \!\left(q_{h,w}^c|\vq_{h,w}^{<c}, \vq_{<(h,w)} \right) ,
\end{aligned}
\end{equation}
where $q^c_{h,w}$ denotes the token at spatial location $(h,w)$ and channel $c$, $\vq^{<c}_{h,w}$ represents tokens at the same position across preceding channels, and $\vq_{<(h,w)}$ encompasses all tokens from prior spatial positions in the generation sequence.

The autoregressive backbone processes preceding spatial positions to provide context features $\vz$ at each location, which serves as an intermediate representation connecting spatial and dimension-wise autoregressive processes. This $\vz$ (referenced in Eq.~\ref{eq:ar_head}) conditions the dimension-wise prediction through our dimension-wise autoregressive head, effectively decoupling the spatial and channel predictions to reduce computational complexity. Our dimension-wise autoregressive head is shared across all spatial positions, adding only a small number of parameters to the model.

\vspace{3pt}
\noindent\textbf{Training.}
During training, we optimize a standard cross-entropy loss applied to the dimension-wise token predictions, enabling simple categorical classification training without requiring complex distribution modeling. 

\vspace{3pt}
\noindent\textbf{Inference.}
At inference time, generation proceeds as follows: (1) the backbone network autoregressively computes context features for each spatial position based on previously generated tokens, (2) for each position, the autoregressive head sequentially predicts values across all channels, and (3) after each spatial token is completely generated, we immediately de-quantize the discrete indices back to continuous features before feeding them into the spatial autoregressive model for the next position's prediction. This de-quantization step is crucial because our autoregressive model takes continuous feature representations as input conditions, ensuring that the network consistently receives features in the original VAE latent space, thus maintaining the rich representational capacity while preserving the advantages of discrete token prediction.
Upon generation completion, all predicted features are decoded into images using the pretrained VAE decoder.
\section{Experiments}
\label{sec:exp}

\begin{figure}[!tbp]
  \centering
  \includegraphics[width=\linewidth]{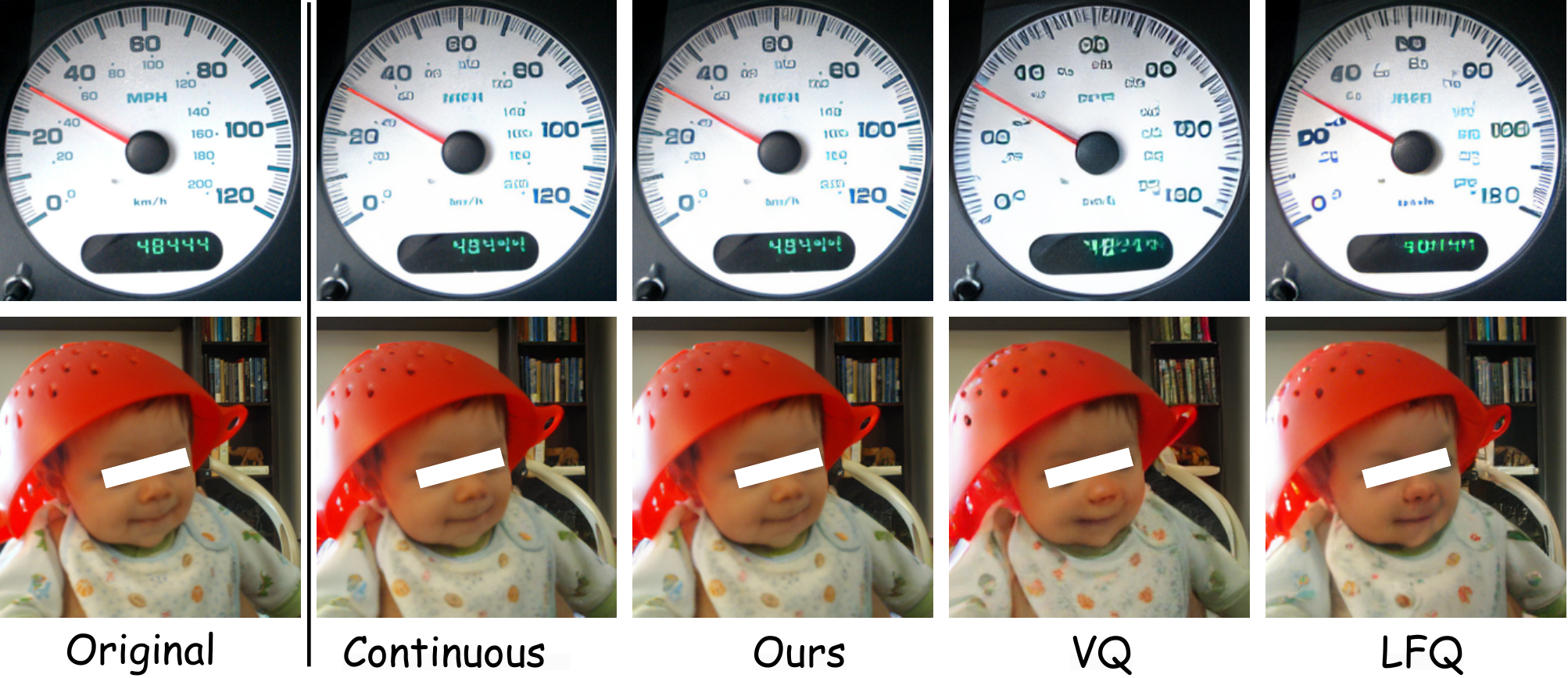}
 \vspace{-20pt}
  \caption{\textbf{Reconstruction quality of typical continuous and  discrete tokenizers.} For discrete baselines, we use VQ from ~\cite{sun2024autoregressive}, and LFQ from ~\cite{openmagvit2}. Our method achieves reconstruction quality comparable to continuous VAE, preserving more fine details than traditional discrete tokenizers, especially in text and facial features. Zoom in for better comparison.}
  \label{fig:token_compare}
   \vspace{-10pt}
\end{figure}

\begin{figure}[!tbp]
  \centering
  \includegraphics[width=\linewidth]{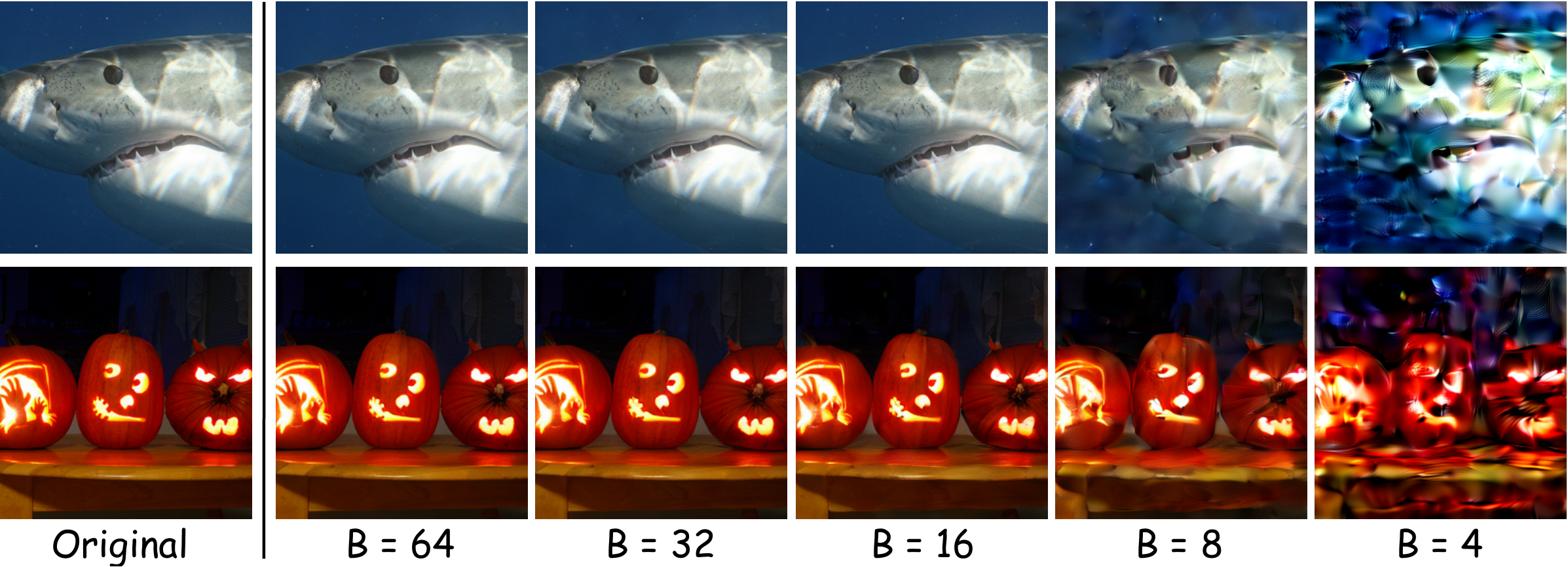}
  \vspace{-20pt}
  \caption{\textbf{Reconstruction quality of different quantization granularities B.} Visual comparison showing reconstructions at decreasing quantization levels. 
  %For $B\geq32$, reconstructions remain visually indistinguishable from the original inputs. As quantization levels decrease, progressive artifacts emerge, though the overall structure is preserved despite significant detail loss. 
  Zoom in for better comparison.}
  \label{fig:quant_level}
   \vspace{-15pt}
\end{figure}

% \begin{figure*}[!tbp]
%   \centering
%   \includegraphics[width=\linewidth]{figs/sota.pdf}
%   \vspace{-10pt}
%   \caption{\textbf{} }
%   \label{fig:sota_compare}
%    \vspace{-5pt}
% \end{figure*}

\subsection{Implementation Details}
\noindent \textbf{Tokenizer.}
For VAE, we use the KL-regularized tokenizer from LDM~\cite{stable_diff} with pre-trained weights from~\cite{mar}. The tokenizer maps 256×256 images to 16×16 tokens with 16-dimensional feature vectors. For dimension-wise quantization, we set $[\alpha_{\text{min}}, \alpha_{\text{max}}]=[-5,5]$, $r=3$, and $B=64$ based on reconstruction quality experiments in Sec.~\ref{sec:4.2}.

\noindent \textbf{Autoregressive Model.}
For a fair comparison with continuous approaches, we adopt the masked autoregressive model architecture from MAR~\cite{mar}. Our default Transformer consists of 32 blocks with a width of 1024 (L model, $\sim$400M) for ablation studies, while final results use a larger H model (with 40 blocks and 1280 width, $\sim$910M).

\noindent \textbf{Evaluation.}
We conduct experiments on ImageNet~\cite{imagenet} at 256×256 resolution. Following common practice~\cite{DDPM_paper}, we evaluate using FID~\cite{fid} and IS~\cite{inception_score}, with Precision and Recall as reference metrics. Default models for ablations are trained for 400 epochs, while final results use 800 epochs. At inference time, temperature ($\tau$) sampling and classifier-free guidance~\cite{ho2022classifier} is applied.

More architecture, training, and inference details can be found in Supplementary Materials.
% \noindent \textbf{Dimension-wise Prediction Head.}
% For each spatial location, our channel-wise prediction head consists of a condition projection layer that maps spatial features to channel embedding space, followed by a stack of transformer blocks with causal attention. During training, we use teacher forcing with the ground truth channel class, and at inference time, we sample each channel autoregressively using temperature $\tau$.

\subsection{Properties of Our Tokenizer}
\label{sec:4.2}

\noindent \textbf{Comparison of Continuous and Discrete Tokenizers.} In Fig.~\ref{fig:token_compare}, we compare the reconstruction quality of typical continuous and discrete tokenizers using  ImageNet~\cite{imagenet} images. 
For continuous tokenization, we use the VAE from~\cite{mar}. For discrete tokenizers, we use VQGAN from LlamaGen's~\cite{sun2024autoregressive} implementation (an improved version of the original~\cite{esser2020taming}), along with  LFQ from Open-MAGVIT2~\cite{openmagvit2}, as the original MAGVIT2~\cite{yu2023language} is not open-sourced. As shown in the figure, continuous tokenizers preserve more details, particularly in text and facial features, where discrete tokenizers often struggle. However, our discrete tokenizer achieves reconstruction quality comparable to its continuous counterpart. %\JS{any finding from the comparison?} 

\begin{table}[t]
\centering
\small
\setlength{\tabcolsep}{3pt}  % 减小列间距，默认是6pt
\begin{subtable}[t]{0.48\linewidth}
\centering
\begin{tabular}{l@{\hspace{4pt}}c@{\hspace{4pt}}|c@{\hspace{4pt}}c}
\toprule
Method & $B$ & rFID$\downarrow$ & IS$\uparrow$ \\
\midrule
VAE & - & 1.11 & 306.1 \\
\midrule
Ours & 8 & 3.69 & 250.3 \\
Ours & 16 & 1.33 & 296.8 \\
Ours & 32 & 1.12 & 303.8 \\
Ours & 64 & \textbf{1.11} & \textbf{305.4} \\
\bottomrule
\end{tabular}
\caption{\textbf{Different quantization levels.} Finer quantization granularity improves performance, with $B$=64 matching continuous VAE.}
\end{subtable}
\hfill
\begin{subtable}[t]{0.48\linewidth}
\centering
\begin{tabular}{l@{\hspace{4pt}}c@{\hspace{4pt}}|c@{\hspace{4pt}}c}
\toprule
Method & Range & rFID$\downarrow$ & IS$\uparrow$ \\
\midrule
VAE & - & 1.11 & 306.1 \\
\midrule
Ours & $\pm{2}$ & 2.26 & 267.9 \\
Ours & $\pm{3}$ & 1.22 & 299.7 \\
Ours & $\pm{4}$ & 1.13 & 305.3 \\
Ours & $\pm{5}$ & \textbf{1.11} & \textbf{305.4} \\
\bottomrule
\end{tabular}
\caption{\textbf{Different quantization ranges.} Feature range $\pm5$ achieves reconstruction quality matching continuous VAE.}
\end{subtable}
\vspace{-10pt}
\caption{\textbf{Ablations on post-training quantization} on the ImageNet Training Set.}
\vspace{-18pt}
\label{tab:tokenizer}
\end{table}

\noindent \textbf{Parameters for Tokenizer Quantization.} 
Tab.~\ref{tab:tokenizer} (a) and Fig.~\ref{fig:quant_level} demonstrate the impact of quantization levels $B$(described in Eq.~\ref{eq:non-uniform-quantization}
) on reconstruction quality. One observation is that global structure remains well-preserved across all quantization levels, with differences primarily in fine detail preservation. With $B=8$, significant information loss occurs (rFID=3.69), visible as artifacts in textures and edges. At $B=16$, quality improves substantially (rFID=1.33), with only minor detail loss visible while only 0.2 rFID higher than the continuous baseline. Reconstructions at $B=32$ and $B=64$ become visually indistinguishable from the original inputs, with $B=64$ (rFID=1.11) matching the continuous VAE baseline perfectly. These results show that our post-training quantization approach can achieve continuous-quality reconstruction with enough quantization levels.

Tab.~\ref{tab:ablations} (b) examines how feature range $[\alpha_{\min}, \alpha_{\max}]$ affects quantization. The approximately Gaussian distribution of VAE features (due to KL regularization) enables defining a finite range. A narrow range of $\pm{2}$ results in information loss (rFID=2.26), while expanding to $\pm{5}$ allows our method to match the continuous baseline (rFID=1.11). Note that to align with MAR~\cite{mar}, the above metrics are evaluated on the training set of ImageNet. For comparison with other works, our best rFID on the validation set is \textbf{0.53}.

\begin{table}[!t]
\centering
% \vspace{-5pt}
% \vspace{-8pt}
\small
\setlength{\tabcolsep}{2pt}
\begin{subtable}[t]{0.48\linewidth}
\centering
\begin{tabular}{l@{\hspace{2pt}}|@{\hspace{2pt}}c@{\hspace{2pt}}c}
\toprule
Prediction & gFID$\downarrow$ & IS$\uparrow$ \\
\midrule
Parallel & 15.7 & 158.5 \\
Autoregressive & \textbf{1.94} & \textbf{306.1} \\
\bottomrule
\end{tabular}
\caption{\textbf{Prediction strategy.} Parallel prediction fails to model inter-dimensional dependencies critical for quality.}
\end{subtable}
\hfill
\begin{subtable}[t]{0.48\linewidth}
\centering
\begin{tabular}{l@{\hspace{2pt}}|@{\hspace{2pt}}c@{\hspace{2pt}}c}
\toprule
Order & gFID$\downarrow$ & IS$\uparrow$ \\
\midrule
Normal & 1.94 & 306.1 \\
Frequency & \textbf{1.89} & \textbf{307.3} \\
\bottomrule
\end{tabular}
\caption{\textbf{Dimension ordering.} Frequency based ordering prioritizes structural information before details.}
\end{subtable}
\begin{subtable}[t]{0.35\linewidth}
\centering
\begin{tabular}{c@{\hspace{2pt}}|@{\hspace{2pt}}c@{\hspace{2pt}}c}
\toprule
B & gFID$\downarrow$ & IS$\uparrow$ \\
\midrule
16 & 2.03 & 295.0 \\
32 & 1.98 & 298.4 \\
64 & \textbf{1.94} & \textbf{306.1} \\
\bottomrule
\end{tabular}
\caption{\textbf{Quantization levels.} Higher granularity consistently improves generation quality.}
\end{subtable}
\hfill
\begin{subtable}[t]{0.58\linewidth}
\centering
\begin{tabular}{c@{\hspace{2pt}}c@{\hspace{2pt}}c@{\hspace{2pt}}|@{\hspace{2pt}}c@{\hspace{2pt}}c}
\toprule
\#ch/g & Classes & \#params & gFID$\downarrow$ & IS$\uparrow$ \\
\midrule
1 & 16 & 60M & \textbf{2.28} & \textbf{289.1} \\
2 & 256 & 63M & 2.45 & 291.2 \\
4 & 65K & 530M & 3.24 & 282.9 \\
\bottomrule
\end{tabular}
\caption{\textbf{Channel grouping.} Joint classification of multiple channels (\#ch/g denotes channels per group) increases parameters yet degrades quality.}
\end{subtable}
\begin{subtable}[t]{\linewidth}
\centering
\begin{tabular}{c@{\hspace{2pt}}c@{\hspace{2pt}}c@{\hspace{2pt}}|@{\hspace{2pt}}c@{\hspace{2pt}}c}
\toprule
Dim & Depth & \#params & gFID$\downarrow$ & IS$\uparrow$ \\
\midrule
256 & 3 & 3M & 2.88 & 277.3 \\
512 & 3 & 10M & 2.72 & 284.8 \\
512 & 4 & 13M & 2.65 & 295.5 \\
1024 & 4 & 65M & 2.03 & 305.0 \\
1024 & 6 & 94M & \textbf{1.94} & \textbf{306.1} \\
\bottomrule
\end{tabular}
\caption{\textbf{Autoregressive head architecture.} Even lightweight design (3M params) achieves reasonable quality. Increasing capacity further enhances performance.}
\end{subtable}
\vspace{-10pt}
\caption{\textbf{Ablation studies on our generation model.}}
\vspace{-18pt}
\end{table}
\label{tab:ablations}

\subsection{Properties of Our Generator}

\begin{table*}[t]
\centering
\small
\adjustbox{width=.80\textwidth,center}{
\begin{tabular}{lllllccccc}
\toprule
Token & Tokenizer & Type & Loss & Method & \#params & FID$\downarrow$ & IS$\uparrow$ & Pre.$\uparrow$ & Rec.$\uparrow$ \\
\midrule
% \multicolumn{9}{l}{\textit{training-quantized discrete tokens}} \\
\multirow{8}{*}{\textit{\shortstack[l]{{training-quantized}\\ {discrete tokens}}}}  & VQ & Mask. & CE & MaskGIT~\cite{chang2022maskgit} & 177M & 6.18 & 182.1 & - & - \\
& VQ & AR & CE & RQTran~\cite{rq} & 3.8B & 7.55 & 134.0 & - & - \\
& VQ & AR & CE & ViT-VQGAN~\cite{vit-vqgan} & 1.7B & 4.17 & 175.1 & - & - \\
& VQ & Mask. & CE & TiTok-128~\cite{yu2024image} & 287M & 1.97 & 281.8 & - & - \\
& LFQ & AR & CE & MAGVIT-v2~\cite{yu2023language} & 307M & 1.78 & 319.4 & - & - \\
& LFQ & AR & CE & Open-MAGVIT2-L~\cite{openmagvit2} & 804M & 2.51 & 271.7 & 0.84 & 0.54 \\
& VQ & AR & CE & LlamaGen~\cite{sun2024autoregressive} & 3.1B & 2.18 & 263.3 & 0.81 & 0.58 \\
& VQ & AR & CE & VAR~\cite{tian2024visual} & 2.0B & 1.73 & 350.2 & 0.82 & 0.60 \\
\midrule
\multirow{8}{*}{\textit{\shortstack[l]{{continuous-valued}\\ {tokens}}}}
% \multicolumn{9}{l}{\textit{continuous-valued tokens}} \\
& VAE & Diff. & Diff. & LDM-4~\cite{stable_diff} & 400M & 3.60 & 247.7 & 0.87 & 0.48 \\
& VAE & Diff. & Diff. & U-ViT-H/2-G~\cite{Bao_2023_CVPR} & 501M & 2.29 & 263.9 & 0.82 & 0.57 \\
& VAE & Diff. & Diff. & DiT-XL/2~\cite{peebles2023scalable} & 675M & 2.27 & 278.2 & 0.83 & 0.57 \\
& VAE & Diff. & Diff. & MDTv2-XL/2~\cite{gao2023masked} & 676M & 1.58 & 314.7 & 0.79 & 0.65 \\
& VAE & AR & GMM & GIVT~\cite{givt} & 304M & 3.35 & - & 0.84 & 0.53 \\
& VAE & AR & Flow & FlowAR-H~\cite{ren2024flowar} & 1.9B & 1.65 & 296.5 & 0.83 & 0.60 \\
& VAE & AR & Diff. & MAR-L~\cite{mar} & 479M & 1.78 & 296.0 & 0.81 & 0.60 \\
& VAE & AR & Diff. & MAR-H~\cite{mar} & 943M & 1.55 & 303.7 & 0.81 & 0.62 \\
\midrule
\multirow{2}{*}{\textit{\shortstack[l]{{post-training quantized}\\ {discrete tokens}}}}
% & \multicolumn{9}{l}{\textit{post-quantized discrete tokens}} \\
    % \rowcolor{gray!20}
& \cellcolor{gray!20} VAE & \cellcolor{gray!20} AR & \cellcolor{gray!20} CE & \cellcolor{gray!20} \textbf{Ours-L } & \cellcolor{gray!20} 486M & \cellcolor{gray!20} 1.76 & \cellcolor{gray!20} 294.8 & \cellcolor{gray!20} 0.80 & \cellcolor{gray!20} 0.63 \\
    % \rowcolor{gray!20}
& \cellcolor{gray!20} VAE & \cellcolor{gray!20} AR & \cellcolor{gray!20} CE & \cellcolor{gray!20} \textbf{Ours-H } & \cellcolor{gray!20} 910M & \cellcolor{gray!20} 1.55 & \cellcolor{gray!20} 313.3 & \cellcolor{gray!20} 0.80 & \cellcolor{gray!20} 0.65 \\
\bottomrule
\end{tabular}}
\vspace{-7pt}
\caption{\textbf{Comparison of visual generation methods on ImageNet 256×256.} Our model achieves comparable performance to the best continuous token approach (MAR) while using standard categorical prediction in autoregressive modeling.}
\vspace{-17pt}
\end{table*}
\label{tab:final}

\begin{figure}[!t]
  \centering
  \includegraphics[width=\linewidth]{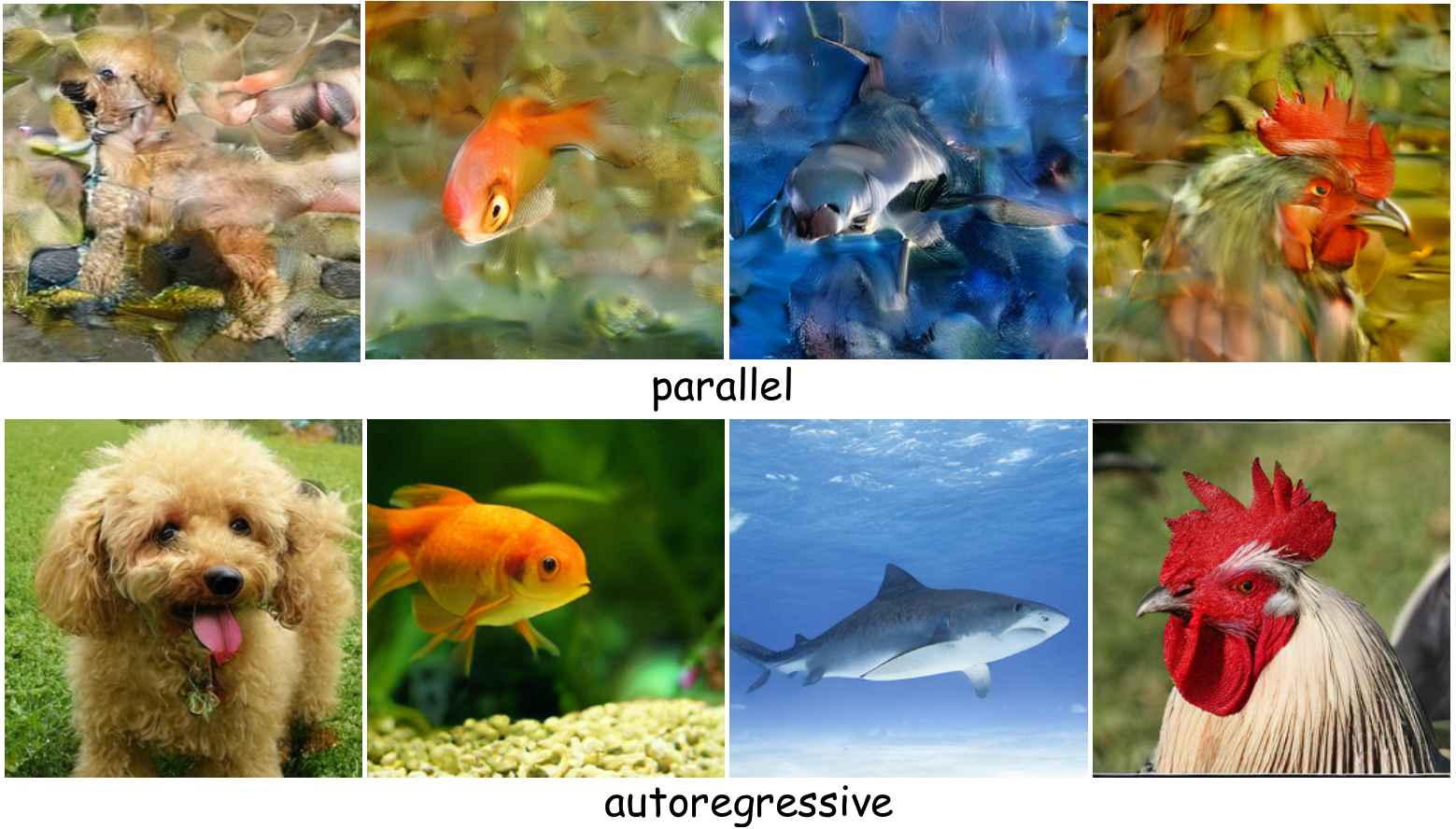}
  \vspace{-20pt}
  \caption{\textbf{Token Prediction Strategy.} Comparison of dimension-wise token prediction approaches. \textbf{Top}: Parallel prediction produces blurry, inconsistent images. \textbf{Bottom}: Our autoregressive approach sequentially predicts token dimensions,  generating coherent, high-quality images. This highlights the interdependence of token  dimensions and they cannot be predicted independently.}
  \label{fig:token_pred}
   \vspace{-6pt}
\end{figure}

\noindent \textbf{Prediction Strategy.} 
Our dimension-wise quantization results in an exponentially large token space. A straightforward solution would be to predict each dimension independently classify each dimension in parallel. However, as shown in the top row of Fig.~\ref{fig:token_pred}, this strategy yields poor results with inconsistent content and blurry artifacts.
As shown in Tab.~\ref{tab:ablations}(a), our dimension-wise autoregressive approach significantly improves generation quality (gFID=1.94, IS=306.1 vs. gFID=15.7, IS=158.5 for parallel prediction). This dramatic 8× improvement in FID confirms that modeling inter-dimensional dependencies is critical for high-quality image generation.

\noindent \textbf{Dimension Ordering.} 
As described in Sec.~\ref{sec:3.3}, the generation order of different dimensions can impact generation quality in our dimension-wise autoregressive framework. As shown in Tab.~\ref{tab:ablations}(b), the frequency-based ordering (prioritizing dimensions with more low-frequency content) slightly improves performance (gFID=1.89 vs. 1.94). While the improvement is modest, it confirms that prioritizing structural information in the generation sequence helps create more coherent images.

\noindent \textbf{Quantization Levels for Generation.}
We investigate how quantization granularity affects generation performance. As shown in Tab.~\ref{tab:ablations}(c), generation quality consistently improves with finer quantization. Even with coarse quantization, our approach achieves reasonable quality (gFID=2.03 at B=16), while finer quantization yields the best results (gFID=1.94 at B=64). This pattern aligns with our reconstruction experiments and confirms that generation benefits from finer-grained discretization.

\begin{figure}[!thp]
  \centering
  \includegraphics[width=\linewidth]{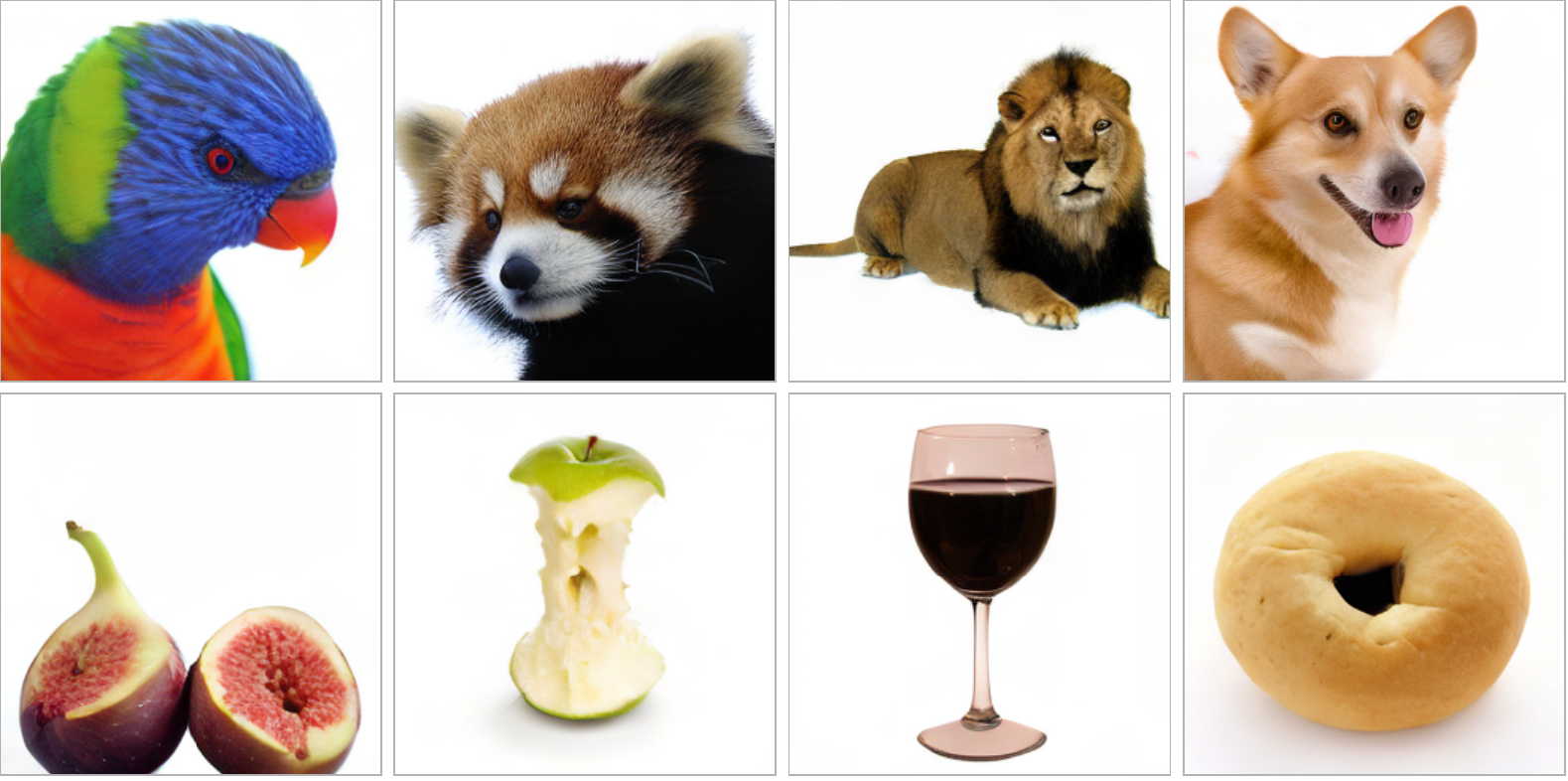}
  \vspace{-20pt}
  \caption{\textbf{Generation guided by token confidence.} Our discrete token approach enables confidence-guided generation, producing clean foreground objects against simple backgrounds by prioritizing high-confidence tokens. This provides a advantage over continuous tokens, which lack explicit token-level confidence scores.}
  \label{fig:token_score}
   \vspace{-17pt}
\end{figure}

\noindent \textbf{Channel Grouping.} We explore whether combining multiple channels for joint classification could improve generation performance. Due to the exponential growth in class count when grouping channels, we conduct experiments with $B$=16 quantization on a 172M parameter backbone. As shown in Tab.~\ref{tab:ablations}(d), our default approach performs classification for each channel. Grouping channels exponentially increases vocabulary size: 2 channels/group requires predicting from 256 classes, while 4 channels/group expands to 65,536 classes. Despite increased parameters (530M vs. 60M for single-channel), generation quality worsens (gFID from 2.28 to 3.24). We suspect this is because larger classification spaces pose greater modeling challenges for the generator, despite the increased parameter count. These results validate the effectiveness of our dimension-wise factorization approach for handling large vocabulary spaces.

\noindent \textbf{Autoregressive Head Architecture.} 
We analyze the impact of autoregressive head capacity on generation quality. As shown in Tab.~\ref{tab:ablations}(e), even our smallest configuration (3M parameters) achieves reasonable quality (gFID=2.88), demonstrating our approach's effectiveness regardless of the scale of the autoregressive head. Increasing capacity consistently improves performance, with our largest configuration (94M parameters) achieving the best results (gFID=1.94).

\noindent \textbf{Confidence-guided Generation.} 
An advantage of our discrete tokenization approach is the availability of explicit token confidence scores during generation. By leveraging these scores, we can implement controlled generation strategies that prioritize high-confidence predictions. As shown in Fig.~\ref{fig:token_score}, selectively generating only the highest-confidence tokens at each step (similar to approaches like MaskGIT~\cite{chang2022maskgit}) results in images with well-defined foreground objects against simplified backgrounds, as the model typically assigns higher confidence to semantically important foreground elements while background details receive lower confidence scores.
Unlike continuous approaches such as MAR~\cite{mar} that lack token-level score measures, our method naturally supports foreground-focused generation. This capability may help with transparent image generation applications (e.g., LayerDiffusion~\cite{zhang2024transparent}) and could potentially facilitate compositional generation where multiple objects are combined into scenes. The ability to distinguish between high and low confidence regions provides finer control over the generation process, representing another advantage of our discrete approach beyond simplified training and inference.

\subsection{Main Results}

Tab.~\ref{tab:final} presents a comparison of our approach against typical visual generation methods on ImageNet-256~\cite{imagenet}. We categorize these methods into three groups: traditional training-quantized discrete token models, continuous-valued token models, and our post-training quantized discrete token models.
Our models demonstrate competitive performance compared to existing approaches. When compared with discrete token methods, Ours achieves better FID scores than most approaches despite their larger model sizes. For instance, LlamaGen~\cite{sun2024autoregressive} with 3.1B parameters yields an FID of 2.18, compared to Ours-L's 1.76 with only 486M parameters.
In comparison with continuous approaches, Ours-L substantially outperforms GIVT~\cite{givt} (FID 1.76 vs 3.35) which uses Gaussian Mixture Models (GMM) for continuous token distribution modeling, and Ours-H achieves better results than FlowAR-H~\cite{ren2024flowar} (FID 1.65) despite the latter having nearly two times the parameter count.
In direct fair comparison with MAR~\cite{mar}, which employs diffusion-based token distribution modeling, Ours-L achieves comparable performance to MAR-L with similar parameter counts. More notably, Ours-H matches MAR-H in FID (1.55) while achieving higher IS and Recall metrics with slightly fewer parameters.
These results confirm that our TokenBridge approach effectively bridges discrete and continuous token representations, achieving high-quality visual generation comparable to continuous methods while maintaining the modeling simplicity of discrete approaches with standard cross-entropy loss.

\section{Conclusion}
\label{sec:conclusion}

In this work, we introduce TokenBridge, a novel approach that bridges discrete and continuous token representations for autoregressive visual generation. We obtain high-quality discrete tokens through post-training quantization while enabling efficient modeling of large vocabulary spaces through dimension-wise autoregressive decomposition. Our work demonstrates that discrete token approaches with standard cross-entropy loss can match state-of-the-art continuous methods without requiring complex distribution modeling techniques. We hope our work will foster future research on high-quality visual generation and unified multimodal frameworks.

\section*{Acknowledgment}
This work is partially supported by the National Nature Science Foundation of China (No. 62402406).
The authors are grateful to Tianhong Li for helpful discussions on MAR and to Yi Jiang, Difan Zou, and Yujin Han for valuable feedback on the early version of this work.

{
    \small
    \bibliographystyle{ieeenat_fullname}
    \bibliography{main}
}
\clearpage
\setcounter{page}{1}
\setcounter{section}{0}
\renewcommand{\thesection}{\Alph{section}}
\maketitlesupplementary

\section*{Appendix}
\addcontentsline{toc}{section}{Appendix}

The supplementary material includes the following additional information:
\begin{itemize}
    \item Sec.~\ref{sec:A} provides implementation details for TokenBridge.
    \item Sec.~\ref{sec:B} presents speed comparison of our token prediction against diffusion-based head.
    \item Sec.~\ref{sec:C} evaluates generalization to different VAE and AE architectures.
    \item Sec.~\ref{sec:D} discusses limitations and broader impacts.
    \item Sec.~\ref{sec:E} showcases additional image generation results.
\end{itemize}

\section{Implementation Details for TokenBridge}
\label{sec:A}

We train our models on the ImageNet-1K~\cite{imagenet} training set, consisting of 1,281,167 images across 1,000 object classes. We adopt the VAE tokenizer from~\cite{mar} and apply our dimension-wise quantization with B=64 levels to its continuous features. For the autoregressive model architecture, we follow MAR~\cite{mar}, with our L model consisting of 32 transformer blocks (width 1024) and H model using 40 blocks (width 1280). Our dimension-wise autoregressive head uses 1024 hidden dimensions with 4 layers for the L model and 6 layers for the H model. At inference time, we employ temperature sampling and classifier-free guidance~\cite{ho2022classifier} to enhance generation quality. The detailed training and sampling hyper-parameters are listed in Tab.~\ref{tab:img_params}.

\begin{table}[h]
\centering
\begin{tabular}{p{0.45\columnwidth}|p{0.45\columnwidth}}
% \toprule
config & value \\
\Xhline{1.2pt}
\multicolumn{2}{c}{\textit{training hyper-params}} \\
\Xhline{0.8pt}
optimizer & AdamW~\cite{loshchilov2017adamw} \\
learning rate & 4e-4 \\
weight decay & 0.02 \\
optimizer momentum & (0.9, 0.95) \\
batch size & 2048 \\
learning rate schedule & cosine decay \\
warmup epochs & 200 \\
ending learning rate & 0 \\
total epochs & 800 \\
dropout rate & 0.1 \\
attn dropout rate & 0.1 \\
class label dropout rate & 0.1 \\
precision & bfloat16 \\
EMA momentum & 0.9999 \\
max grad norm & 1.0 \\
\Xhline{0.8pt}
\multicolumn{2}{c}{\textit{sampling hyper-params}} \\
\Xhline{0.8pt}
temperature & 0.97(L) / 0.91(H) \\
CFG class dropout rate & 0.1 \\
guidance scale & 3.1 (L) / 3.45 (H) \\
\Xhline{0.8pt}
\end{tabular}
\caption{\textbf{Detailed hyper-parameters for TokenBridge.}}
\label{tab:img_params}
\end{table}

\section{Speed Comparison of Token prediction}
\label{sec:B}

We compare the speed of our dimension-wise prediction approach with MAR's~\cite{mar} diffusion-based approach. Table~\ref{tab:speed_comparison} shows the results.

\begin{table}[h]
\centering
\begin{tabular}{l|c}
\Xhline{1.2pt}
Method & Time (ms) \\
\Xhline{0.8pt}
Diffusion (MAR) & 311.25 ± 1.85 \\
AR (Ours) & 52.42 ± 0.57 \\
\Xhline{1.2pt}
\end{tabular}
\caption{\textbf{Comparison of single image token prediction time. }All measurements conducted with batch size 1 on an NVIDIA A100 GPU, averaged over 100 runs. Our method achieves a 5.94× speedup over MAR's diffusion sampling (100 steps).}
\label{tab:speed_comparison}
\end{table}

As shown in Table~\ref{tab:speed_comparison}, our approach is 5.94× faster than MAR's~\cite{mar} diffusion-based~\cite{DDIM_paper} token prediction. This efficiency advantage comes from our dimension-wise autoregressive prediction strategy that directly generates discrete tokens without iterative sampling procedures. Although our method requires sequential prediction steps (one per channel), the lightweight design of our AR head and the ability to utilize KV cache in transformers maintain high efficiency compared to diffusion sampling.

The number of prediction steps in TokenBridge corresponds to the VAE~\cite{vae} channel count (16 in our implementation). With newer architectures like SDXL's~\cite{podell2023sdxl} VAE that use only 4 channels, our approach would require even fewer steps. 

\section{Generalization to Different VAE and AE Architectures}
\label{sec:C}

\begin{figure}[h]
\centering
\includegraphics[width=.8\columnwidth]{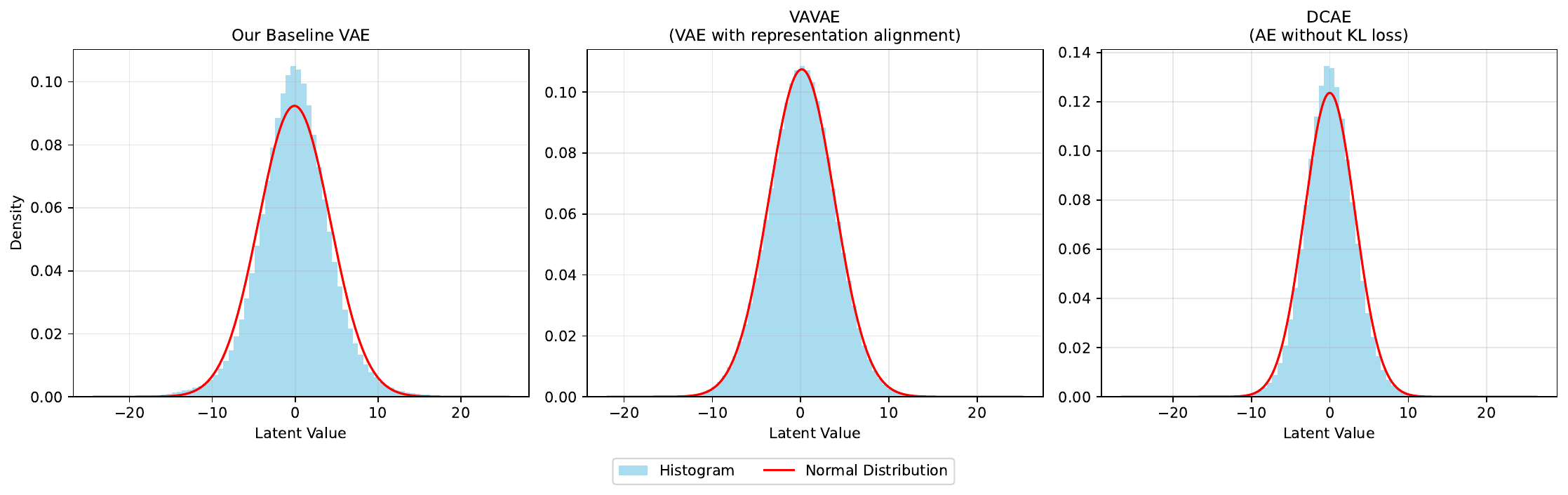}
\caption{\textbf{Latent distributions of different autoencoders.} Despite architectural differences and training objectives, all three models exhibit similar near-Gaussian distributions, validating the generalizability of our quantization approach.
}
\label{fig:vae_vis}
\vspace{-15pt}
\end{figure}

To evaluate the generalization of our post-training quantization approach, we select two representative alternative autoencoders for evaluation: VAVAE~\cite{vavae}, a state-of-the-art VAE with representation alignment, and DCAE~\cite{dcae}, achieving high compression rates without KL loss constraints.
Fig.~\ref{fig:vae_vis} visualizes the latent feature distributions of different autoencoders. Although the value ranges differ across architectures, all exhibit similar near-Gaussian distributions. This consistency validates that the bounded, approximately Gaussian property, independent of specific architectural designs or training constraints like KL regularization. As described in the Method section, even linear quantization achieves good reconstruction results, demonstrating the robustness of our approach across different quantization schemes. 

Tab.~\ref{tab:recon_fid} shows reconstruction results after applying our quantization with corresponding rescaling and quantization granularity. Our method successfully preserves reconstruction quality across different architectures: VAVAE achieves identical performance (rFID=0.28) to its continuous baseline using B=128 and r=3.5, while DCAE matches its baseline (rFID=0.77) with B=64 and r=8. These results demonstrate that our post-training quantization approach generalizes effectively across diverse autoencoder architectures while maintaining reconstruction fidelity.

\begin{table}[h]
\centering
\footnotesize
\setlength{\tabcolsep}{3pt}
\begin{tabular}{l|c|c|c|c}
\toprule
AE & Ori. FID & B & Range & TokenBridge FID \\
\midrule
VAVAE & 0.28 & 128 & [-3.5, 3.5] & 0.28 \\
DCAE & 0.77 & 64 & [-8, 8] & 0.77 \\
\bottomrule
\end{tabular}
\caption{\textbf{Reconstruction quality across different autoencoder architectures.} Our post-training quantization preserves reconstruction fidelity when applied with appropriate parameters.}
\label{tab:recon_fid}
\end{table}

\section{Limitations and Broader Impacts}
\label{sec:D}
\noindent\textbf{Limitations.} 
Our approach inherits limitations from the underlying VAE~\cite{vae} model. The representation quality of the pretrained VAE directly affects our reconstruction fidelity and generation capabilities. We note that further improvements in continuous tokenizer would directly benefit our approach.

\noindent\textbf{Broader Impacts.} 
Our work demonstrates that standard autoregressive modeling with cross-entropy loss can achieve quality comparable to more complex approaches. This finding may encourage simpler model designs in visual generation tasks and facilitate unified multimodal modeling based on autoregressive frameworks. Like all generative models, TokenBridge may reflect biases present in training data and could potentially be misused to create misleading content, which warrants careful consideration in deployment.

\section{More Visualization Results}
\label{sec:E}
\begin{figure*}[!tp]
  \centering
   \vspace{10pt}
  \includegraphics[width=\linewidth]{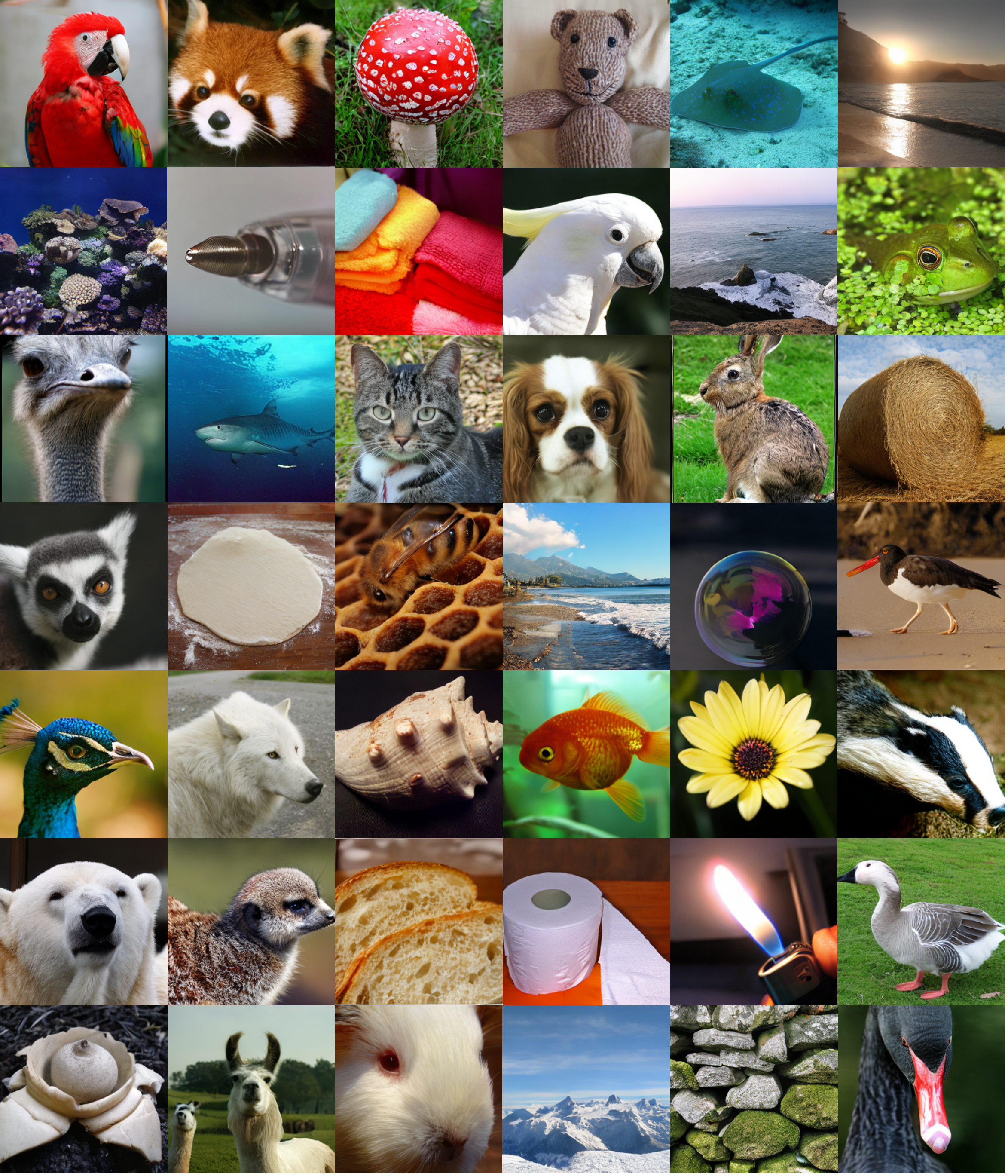}
  % \vspace{-16pt}
  \caption{\textbf{Additional image generation results of TokenBridge across different ImageNet~\cite{imagenet} categories.}
  }
  \label{fig:supp_vis_16}
   % \vspace{-16pt}
\end{figure*}

\end{document}